\documentclass[a4paper,12pt]{article}
\usepackage[dvipsnames]{xcolor}
\usepackage{lscape}
\usepackage{longtable}
\usepackage{authblk}
\usepackage{hyperref}
\usepackage{arabtex}
\usepackage{utf8}
\usepackage[utf8]{inputenc}
\usepackage{amsfonts} 
\usepackage{float}
\setcode{utf8}
\usepackage{enumitem}
\usepackage{CJKutf8}
\usepackage{graphicx}
\usepackage{multirow}

\usepackage[a4paper,top=3cm,bottom=2cm,left=3cm,right=3cm,marginparwidth=1.75cm]{geometry}

\title{Towards a responsible machine learning approach to identify forced labor in fisheries}

\author[1,*]{\small Roc\'io Joo}
\author[2,3,4]{\small Gavin McDonald}
\author[1]{\small Nathan Miller}
\author[1]{\small David Kroodsma}
\author[1]{\small Courtney Farthing}
\author[5]{\small Dyhia Belhabib}
\author[1]{\small Timothy Hochberg}

\affil[1]{\footnotesize Global Fishing Watch, Washington, DC 20036, USA}
\affil[2]{\footnotesize Marine Science Institute, University of California, Santa Barbara, Santa Barbara, CA, USA}
\affil[3]{\footnotesize Bren School of Environmental Science \& Management, University of California, Santa Barbara, Santa Barbara, CA, USA}
\affil[4]{\footnotesize Environmental Markets Lab, University of California, Santa Barbara, Santa Barbara, CA, USA}
\affil[5]{\footnotesize Nautical Crime Investigation Services, Vancouver, BC, Canada}
\affil[*]{Corresponding author: Roc\'io Joo, rocio.joo@globalfishingwatch.org}

\date{}

\begin{document}

\maketitle

\begin{abstract}

Many fishing vessels use forced labor, but identifying vessels that engage in this practice is challenging because few are regularly inspected. We developed a positive-unlabeled learning algorithm using vessel characteristics and movement patterns to estimate an upper bound of the number of positive cases of forced labor, with the goal of helping make accurate, responsible, and fair decisions. 89\% of the reported cases of forced labor were correctly classified as positive (recall) while 98\% of the vessels certified as having decent working conditions were correctly classified as negative. The recall was high for vessels from different regions using different gears, except for trawlers. We found that as much as $\sim$28\% of vessels may operate using forced labor, with the fraction much higher in squid jiggers and longlines. This model could inform risk-based port inspections as part of a broader monitoring, control, and surveillance regime to reduce forced labor.
 * Translated versions of the English title and abstract are available in five languages in S1 Text: Spanish, French, Simplified Chinese, Traditional Chinese, and Indonesian.
\end{abstract}


\section*{Teaser}
Machine learning to identify risk of forced labor in fisheries and assessment of its accuracy, fairness, and usefulness.

\newpage

\section*{Introduction}

The use of forced labor in the fishing industry is a severe problem, documented by numerous investigations, reports, and scientific articles \cite{Marschke2016, Simmons2014, Stringer2016, Vandergeest2021a, Yen2021}; however, the true extent of this practice remains unknown. 
Forced labor has been defined by the International Labour Organization (ILO) as all work or service which is exacted from any person under threat of a penalty and for which the person has not offered themselves voluntarily, and 11 indicators were established by the ILO to help identify the problem: abuse of vulnerability, deception, restriction of movement, isolation, physical and sexual violence, intimidation and threats, retention of identity documents, withholding of wages, debt bondage, abusive working and living conditions, and excessive overtime \cite{SAP-FL2012}. 
In 2007, the 96th Session of the ILO Conference adopted the Work in Fishing Convention (C188), which aims to ensure that fishers have decent work conditions measured against a set of minimum requirements for work on board fishing vessels; conditions of service; accommodation and food; occupational safety and health protection; medical care and social security. 
This treaty has been ratified by only 20 States \cite{ILOC188} and is not yet widely implemented. Other measures related to addressing forced labor in commercial fisheries have been developed regionally and nationally, but again, implementation lags behind. 
Until there is comprehensive ratification and implementation of C188, regional and national labor-related measures, inspections remain few and far between. Without consistent data collection, data sharing when labor violations are identified is almost non-existent.
Even within nations that have ratified the treaty, there can be limited capacity to inspect vessels.
Plus, mechanisms for victims to share their experiences or report their circumstances are not sufficient, meaning that complaints are limited and legal evidence can be hard to find \cite{Ridings2021}. 
It is thus impossible to know for certain how commonly forced labor occurs, where, and on which vessels.
In the absence of exhaustive data, a model can help fill in the blanks by learning to recognize patterns relating to forced labor, with the assumption that vessels that have forced labor behave differently from those that do not---potentially fishing longer hours, staying at sea longer, visiting different ports, meeting up with specific types of vessels at sea, manipulating their transponders, or other types of observable activity. 
Such a model can find vessels that behave similarly to those with forced labor, thus prioritizing inspections at port or in coastguard operations.

The structure and performance of any pattern recognition or classification model rely heavily on the characteristics and amount of data available to train the model with. 
Several data challenges regarding forced labor were highlighted in \cite{mcdonald2021reply}.
Two primary concerns are that the number of reported cases publicly available is small and that the sample is not random nor probabilistic, but rather convenient and likely driven---and biased---by the interests and contacts of journalists, NGOs, and other sources of information. 
In addition, it may be difficult to determine over what time period a vessel used forced labor as the abused crew may not be able to determine the exact dates when they were onboard a particular vessel, especially when they may have worked aboard several vessels in succession or been transshipped between vessels at sea. 
Moreover, a lack of inspection and certification data makes it difficult to identify non-offender cases to train the model. 
Though some human rights and social responsibility certifications exist, they are usually based on one-time inspections while the vessel is in port and do not guarantee that a fishing vessel does not engage in forced labor practices while at sea.
Lastly, information related to the legal protection of workers, such as health inspections on board, crew number and composition, visa requirements in the flag State of the vessel, and presence of worker's unions in the fishery, among others, are not available for most fishing vessels. 

A first approach to identifying forced labor at a global scale while dealing with these constraints was presented in \cite{McDonald2021}. 
Because of the lack of precision in the dates of forced labor events, the authors worked at the scale of a year, i.e. the model trained and predicted on fishing vessel data corresponding to a calendar year, 
aiming to classify vessel-years into positive (forced labor likely occurred) or negative (forced labor did not occur). 
They used a positive-unlabeled (PU) machine learning algorithm training on 21 positive and 66,336 unlabeled vessel-years.
The features used for classification corresponded to vessel characteristics and movement patterns computed from Automatic Identification System (AIS) data.
McDonald's et al \cite{McDonald2021} model correctly identified between $92\%$ and $100\%$ of vessel-years as positive and estimated that between $10\%$ and $20\%$ of the unlabeled vessel-years would be positive. 
This initial proof-of-concept had several technical problems.
First, only 21 positive cases were available to train and validate the model.
Second, there was no data available to assess the model's capability to correctly classify negative vessel-years.
Third, the variability in the predictions and model performance was generated by changing the time scale of the vessel data (vessel-year, vessel-two-years, etc.) which should not be considered a source of model variability but viewed as a complete change in the scale of the study and in the definition of the individual cases of labor abuse. 
Fourth, the hyperparameters of their classifiers--including the random forests, which showed the best performance--were not calibrated and default values were used.
Fifth, to convert positive scores (ranging from 0 to 1) into `positive' and `negative' labels, the threshold that maximized a modified F1 score was used. 
Such a threshold is biased in PU learning scenarios \cite{Menon2015}, resulting in biased scores and classifications. 
Sixth, final predictions by gear and flag State were shown, without any fairness checks to ensure that predictions were not biased against certain gears or flag States. 
A recent study \cite{Karthikeyan2022} used a similar PU learning approach comparing several algorithms, concluding that an artificial neural network performed best. 
However, neither the scale of the study (e.g. vessel-year), the number of positive cases, nor the features used to train the model were described. 
There was no method to account for variability or uncertainty in the predictions, and lacked a detailed description of the hyperparameters. 
Instead of a modified F1-score as in \cite{McDonald2021}, a 0.5 threshold was used for classification. 
Finally, as in \cite{McDonald2021}, there were no fairness checks. 

In this work, we used the modeling approach described in \cite{McDonald2021} as a starting point and modified it to address the aforementioned problems.
We built a positive training set of 72 vessel-years, increasing the positive training data in \cite{McDonald2021} by more than a factor of three.
We also obtained a list of vessels that were inspected by government authorities and certified as compliant with the ILO C188 as not using forced labor and assessed if the model classified them as such (negative).
Internal variability of the algorithm was accounted for by using several initial random seeds at different modeling stages, and the results from different seeds were used to compute a level of confidence in each prediction.
A sensitivity analysis and calibration procedure was performed in order to choose the hyperparameter values in the model classifiers and the number of initial seeds. 
Instead of using a modified F1-score, we used a statistical density-based approach \cite{Ivanov2020} that estimates an upper bound of the true number of positives within the unlabeled cases to calibrate a classification threshold. 
Using an upper bound is necessary as the true proportion of positives is unidentifiable without strong assumptions on the distributions of positives and negatives.
This might result in an overestimation of the number of cases of forced labor (positives), 
though, in practice, considering the lack of understanding of the scale of the problem, and the general under-monitoring of forced labor, 
the downsides of an overestimation is small if the model is not used as evidence in court but rather as a guidance for (more) inspections. 
We assessed the model's fairness as its capacity to correctly identify reported positive cases by fishing gear and flag region. 
In light of our results and based on methodological rigor and transparency criteria, we discuss if the model and its predictions can be trusted for use by competent authorities, directions for improvement, and potential avenues of research for the model to help improve port inspections.

\section*{Materials and Methods}

\subsection*{Reported cases of forced labor}

358 reported cases of vessel-level forced labor were collected. 
They were extracted from gray literature reports, media sources and government reports in 11 languages (mainly English, but also Indonesian, Spanish, Chinese, Khmer, Filipino, French, German, Norwegian, Thai, and Vietnamese), found online using combinations of keywords such as `vessel', `boat', `forced labour', `slavery', `poor living conditions', `death of crew', `murder', or `buried at sea' along with the keywords associated with the ILO Indicators of Forced Labour, and enforcement keywords such as `arrest' or `caught'. 
We also obtained the reported cases used in \cite{McDonald2021} and others from organizations such as the Environmental Justice Foundation, Sea Shepherd, C4ADS, and Human Rights at Sea.
To be included, the cases had to contain witnesses or testimonies from victims, information on prosecutions, arrests, or detentions of vessels under suspicion of forced labor behavior alluding to at least one of the 11 ILO Forced Labor Indicators. 

Though the data collection was as exhaustive as possible, the cases came from multiple sources that are unlikely to have performed any probabilistic sampling design to get representative samples; 
rather they were biased by the networks the reporters and organizations had, and were potentially focused on specific fisheries where global attention is focused.
Moreover, the choice of words and languages for the online search by our experts could have introduced an additional layer of bias to the data.

Vessels were reported to use forced labor for varying time periods, and in many reports, the duration was not clear, with only the date of vessel departure from the port or the date of vessel arrival at the end of a trip reported. 
In some cases, fishers were transferred from one ship to another at sea, and the time of transshipment was not available. 
For this reason, we worked at the scale of a year, and in our data processing, we considered a single report documenting several years of forced labor abuse by a vessel as multiple yearly reports (i.e. multiple positive vessel-years) involving the same vessel and the same source of information.

\subsection*{AIS data}

Automatic Identification System (AIS) data, obtained from satellite providers ORBCOMM and Spire, were processed using Global Fishing Watch’s data pipeline \cite{Kroodsma2018}. 
The pipeline includes machine learning algorithms that classify vessels into specific types (e.g. longliner, purse seiners, squid jigger, trawler) and estimate the vessels’ size, including engine power, length, and gross tonnage, based on movement patterns (combining the results with vessel registries when available). For identified fishing vessels, a second algorithm determines when these vessels are fishing.
GFW's dataset captures the majority of activity by large ($>24$ meters) industrial fishing fleets \cite{Taconet2019}. AIS position data can be directly used to determine spatial information such as distance to port or shore, whether or not  the activity is within an Exclusive Economic Zone (EEZ) or on the high seas, as well as distance traveled. GFW's pipeline also produces a number of derivative products that we could leverage, including port visits, voyage duration between ports, suspected transshipment events between two vessels, loitering events, and gaps in AIS transmission.

\subsection*{Preprocessing}

We matched the reported cases with AIS data using the Maritime Mobile Service Identity (MMSI) of the vessel. In cases when vessel MMSI was not available, the International Maritime Organization (IMO) number was used. 
If there was no match using MMSI or IMO number, we attempted a match using call sign if available. 
If there was still no match, we used ship name, except in cases where the ship name was commonly used by multiple vessels, which made the matches unreliable. From the collected 358 reported cases, only 72 were associated with AIS data for the reported year and used in this work. 

These 72 vessel-year individuals---referred to as positive cases---consisted of 40 drifting longliners, 14 squid jiggers, 11 trawlers, and 7 purse seiners, and flagged to China (34), the Fishing entity of Taiwan (14), Vanuatu (6), Republic of Korea (5), Japan (3), Fiji (2), France (1), Faroe Islands (1), Ireland (1), Malaysia (1), Portugal (1), Russian Federation (1), Thailand (1), and United States of America (1). 
The unlabeled set, i.e. vessel-years with AIS data that did not match reported cases but we could not confirm that were free from labor abuse, was composed of 107,256 vessel-years (see Table \ref{table:groups} for numbers by gear and flag region). 

In addition, we had yearly AIS data from 53 trawlers that had been inspected by government authorities and certified as meeting the criteria in the ILO C188.
This data was obtained through confidentiality agreements so we cannot disclose vessel characteristics. 
AIS data from the year of certification corresponded to the vessel-year of these `negative' cases. 
However, we cannot guarantee that no labor abuses were committed while they were at sea for the duration of the certification. 
For that reason, we did not use the negative cases to train the model, but rather for `expert-judgment-based validation' to compute specificity, described in the section `Assessing accuracy and fairness'.
We removed from the unlabeled set vessels that were reported as positive in other years---thus suspected to be positive in those years---or had been certified as free from labor abuses in other years---thus suspected to be negative in those years.
Table \ref{table:sets} summarizes each of the described datasets.

\begin{table}[ht]
  \caption{Datasets of vessel-year.}
  \label{table:sets}
  \centering
  \begin{tabular}{lllr}
    \hline
    Dataset & Description & Role & Number of \\
    & & & vessel-years \\
    \hline
    Positive cases & 
    Vessel-years from reported & Training,  testing, & 72 \\
    & labor abuses with reliable &  and computation of & \\
    &  AIS data. & performance metrics. & \\
    Negative cases & Vessel-years from certifications  &
    Predictions and & 53 \\
    & with reliable AIS data. &  post hoc comparisons. & \\
    Unlabeled cases & Vessel-years that were not & Training, & 107,256 \\ 
    & known to be positive & testing and & \\
    & or negative. & predictions & \\
    \hline
  \end{tabular}
\end{table}

We used the same features as in \cite{McDonald2021} with very few modifications that accounted for recent changes and improvements in GFW's algorithms. Whereas the previous analysis could only leverage data from 2012-2018, we now leverage more recent data from 2012-2020. The following features were computed for each vessel-year as potential predictors: AIS device type (A or B), average distance (in a straight line projection) between the location where an AIS gap in transmission of at least 12 hours begins and the location where the gap ends, average distance from port (km), average distance from shore (km), average encounter duration (hours), average gap length for gaps in AIS transmission of at least 12 hours (in days), average loitering duration (hours), average number of daily fishing hours, average voyage duration in hours, engine power (kW), fishing gear type (drifting longline, squid jigger, trawler, or purse seiner), flag of convenience (yes or no), maximum distance from port (km), maximum distance from shore (km), number of AIS position messages, number of encounters with other vessels at sea, number of encounters with vessels known to have used forced labor, number of fishing hours, number of fishing hours in foreign EEZs, number of fishing hours in the High Seas, number of foreign port visits, number of gaps (between consecutive AIS messages) of at least 12 hours, number of hours at sea, number of loitering events (potential encounters at sea in which one vessel may not be operating with AIS), number of port-of-convenience port visits, number of voyages, tonnage (gross registered tonnage), total distance traveled (km), and vessel length (meters).

\subsection*{The model}

\begin{figure}[!ht]
    \includegraphics[width=\textwidth]{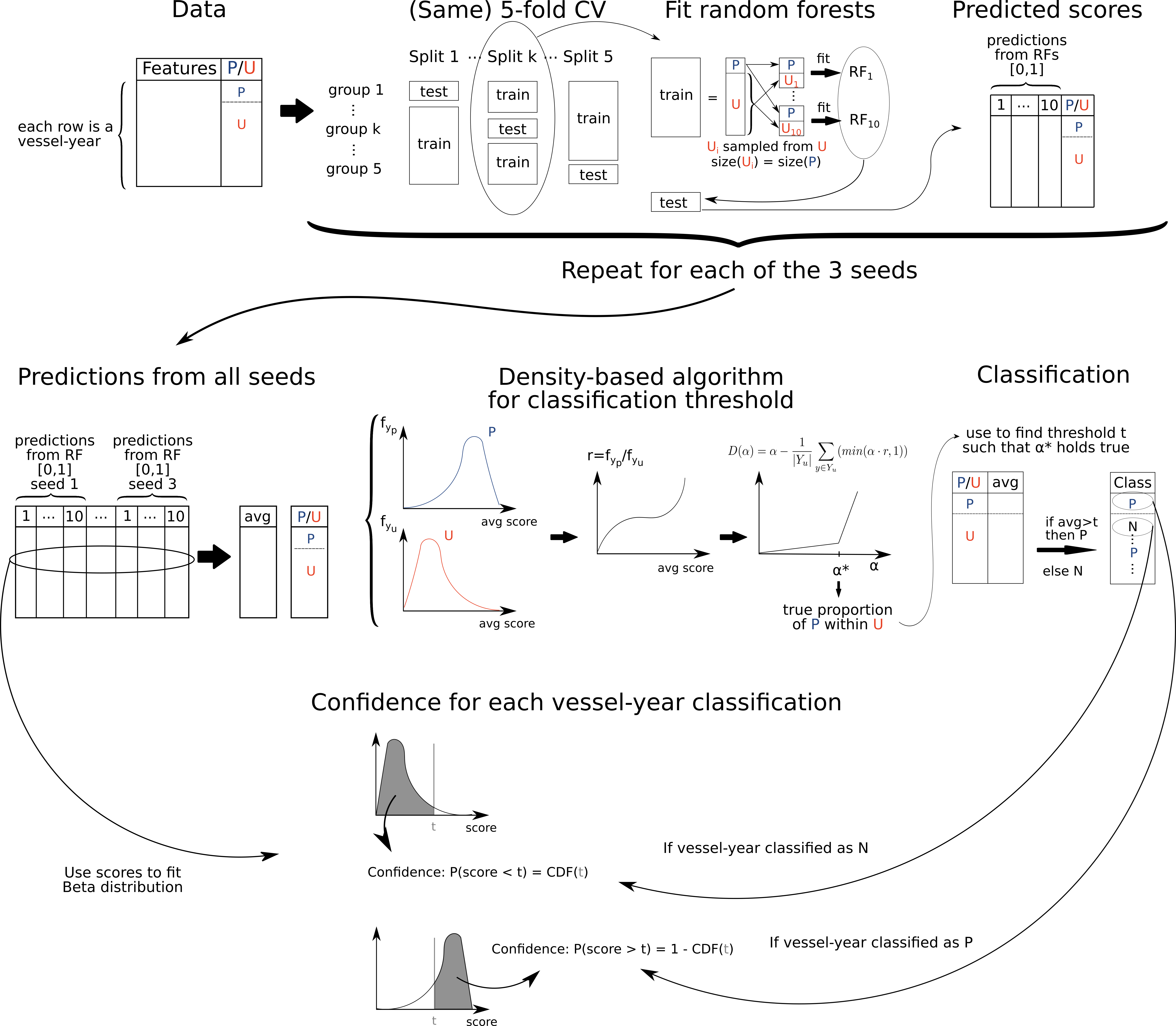}
    \caption{Modeling workflow.}
    \label{fig:workflow}
\end{figure}

The model is summarized in Fig. \ref{fig:workflow}. Typically, to train supervised classification models to identify two categories (e.g. positive vs. negative forced labor), one would provide both labels. 
However, because we could not reliably obtain sufficient, high-confidence examples of vessels without forced labor, we implemented a Positive-Unlabeled (PU) learning \cite{Bekker2020} approach using the positive and unlabeled cases for training and testing. 
The following steps describe our PU learning approach with random forest classifiers (RF; \cite{Breiman2001}):
\begin{enumerate}
\item Five-fold cross-validation split: 
The data (positives and unlabeled) were split into five groups of roughly equal size. 
One of those groups was designated as the test set and the others were to be part of the training set.  
The role of test set was rotated among the groups (five times).
To reduce the risk of data leaking, we kept data from the same source (e.g. media article or report) in the same group, i.e. either all in training or all in test. For instance, if two vessels were mentioned in the same forced labor report they were likely to show similar movement patterns at the same time and be related to each other, so when using one of them in the training set, the model would likely perform well on the other one, an artificial effect of using related cases to train and test. 
\item Training stage and prediction scores: Using the cross-validation sets, we trained RFs and predicted over the test set and the negative cases.
To reduce the weight of the unlabeled cases in the model, we randomly downsampled them in the training set with a 1-1 ratio, i.e. the number of positive and unlabeled cases used for training would be equal. 
The downsampling procedure was done 10 times (i.e., 10 bags) for each of the five splits.
For each sample, the algorithm worked as follows:
\begin{enumerate}
    \item Compute the variance of the (potential) numerical predictors and Pearson correlations between them. 
    Only keep those predictors with variances computationally larger than zero and bivariate correlations smaller than 0.75. Feature selection under these conditions was done using the \texttt{recipes} R package \cite{Rrecipes}.
    \item Fit the RF to the sample set.
\end{enumerate}
\item In each split, use the ten RFs fitted to the training sets to predict on the test set and the negative cases. The prediction outputs are scores from 0 to 1 to classify each case as positive.
\item Repeat steps $1$-$3$: They were performed using a fixed initial random seed to generate the groups for cross-validation and for downsampling. We repeated the three steps for three different initial seeds, leading to 30 RF scores for each case. 
\item Average scores: 
We took average scores per case.
\item Calibrate a classification threshold:
The standard approach in classification problems is to assume that a 0.5 score is enough to classify an individual as positive.
Other approaches estimate a classification threshold by testing different values and choosing the one that optimizes a given performance metric (e.g. the modified F1-score in \cite{McDonald2021}).  
However, the latter assumes that the unlabeled cases are all negative, resulting in biased scores and classifications. 
\cite{Ivanov2020} showed that while the true proportion of positive cases in a population of unlabeled individuals ($\alpha$) is generally unidentifiable \cite{Blanchard2010}, it is possible to estimate an upper bound of this proportion ($\alpha^*$). 
Let's denote $f_{x_p}(x)$ and $f_{x_n}(x)$ the probability density functions for vectors of features $x$ that correspond to positive and negative. 
$x$ in the unlabeled space can be assumed to arise from a mixture distribution with density
\begin{equation}
\label{eq1}
    f_{x_u}(x) \equiv \alpha f_{x_p}(x) + (1-\alpha) f_{x_n}(x)
\end{equation}
where $\alpha \geq 0$ and $(1-\alpha) \geq 0$ are the mixing proportions. 
In this set-up, the probability of being positive given the observed features $x$, $p_p(x)$, also called posterior probability, can be defined as:
    \begin{equation}
    p_p(x) \equiv \frac{\alpha f_{x_p}(x)}{\alpha f_{x_p}(x) + (1 - \alpha)f_{x_n}(x)} = \alpha \frac{f_{x_p}(x)}{f_{x_u}(x)}
    \end{equation}
According to \cite{Ivanov2020}, the corresponding posterior $p^{*}_p(x)$ to $\alpha^*$ can be expressed as a function of the densities of the scores $y$ obtained by a classifier; i.e.:
\begin{equation}
    p^{*}_p(x) \equiv \alpha^* \frac{f_{x_p}(x)}{f_{x_u}(x)} = \alpha^* \frac{f_{y_p}(y)}{f_{y_u}(y)} \equiv p^{*}_p(y)
\end{equation}
And, with $D(\alpha) = \alpha - \mathbb{E}(p(Y_u))$, where $\mathbb{E}(p(Y_u))$ can be estimated as $\hat{\mathbb{E}}(p(Y_u)) = \frac{1}{|Y_u|}\sum_{y \in Y_u}(\min (\alpha \frac{f_{y_p}(Y_u)}{f_{y_u}(Y_u)}, 1))$, $D$ is zero below $\alpha^*$ and increases monotonically above it. 
However, this is only approximately true in practice, so finding $\alpha^*$ comes down to finding the upward bend in $D$. 
We adapted Ivanov's algorithm to estimate $\alpha^*$ as follows:
\begin{enumerate}
    \item Density kernels are estimated for positive and unlabeled average scores ($f_{y_p}$ and $f_{y_u}$, respectively).
    \item For each of the unlabeled average scores $Y_u$, the kernels are used to infer the probability of being positive or unlabeled ($f_{y_p}(Y_u)$ and $f_{y_u}(Y_u)$, respectively).
    \item Next, a density ratio $r=\frac{f_{y_p}(Y_u)}{f_{y_u}(Y_u)}$ is computed, then monotonized (i.e. partial monotonicity of $r$ is enforced) and smoothed with a rolling median.
    \item We then approximate $\alpha^*$ as the value that maximizes the second derivative of $D$ ($\alpha^* = \displaystyle\arg\max_{\alpha} \frac{d^2(D)}{d(\alpha^2)}$), using the definitions above. 
\end{enumerate}
The threshold of classification $t$ is such that the proportion of positive predictions is equal (or approximately equal) to $\alpha^*$.
\item Classify scores into positive or negative: 
We used $t$ to classify the average confidence scores into positive or negative. 
\item Computing confidence for each classification: For each case, we used all 30 scores to fit a Beta distribution. If the case was classified as positive, the confidence in that classification would be equal to the probability of having a score greater than $t$, i.e. $P(y)>t$. 
Conversely, if the case was classified as negative, the confidence in that classification would be equal to the probability of having a score lesser than $t$, i.e. $P(y)<t$.
\end{enumerate}
This algorithm is referred to in the other sections as `the model'. The number of bags (10), the values of RF hyperparameters, and initial random seeds (three) were fixed after sensitivity analyses and calibration procedures aiming to maximize stability and accuracy while minimizing model complexity (see Supp. Mat. for more details).

\subsection*{Assessing accuracy and fairness}

The main interest in machine learning algorithms is their capacity to produce accurate results. 
Since the goal of our model is to correctly classify vessel-years into offenders or not (i.e. positive or negative) and make global assessments of labor abuses at sea, the overall performance is evaluated through recall, or the proportion of positive cases predicted by the model as positive, and specificity, or the proportion of negative cases predicted by the model as negatives. 

Statistical non-discrimination or fairness measures aim to assess the absence or degree of discrimination of given groups regarding classification \cite{barocas-hardt-narayanan}.
In the context of this work, we evaluated fairness by assessing if the model was equally successful at classifying vessel-years as positives when considering different fishing gears and different flag regions.
We know that fishing gear might be one of the most important factors in conditioning spatial behavior at sea. 
However, there was an unequal representation of gear types in the positive and the unlabeled sets (Table \ref{table:groups}).
While the broader global dataset is dominated by trawlers, the positive cases are dominated by drifting longliners. 
This inequality can make the model more sensitive to the few cases of the underrepresented vessel classes in the positive set.

On the other hand, the flag State of the vessel could be a sensitive characteristic and potentially associated with discrimination.
The majority of the positive cases involved Asian-flagged vessels (Table \ref{table:groups}) so it is possible that the model results could be biased against vessels from that region, with the model able to identify Asian-flagged vessels in reported cases as positive correctly, but less able to identify non-Asian flagged vessels in reported cases as positive.
There was not enough data to compute fairness metrics by flag State, but we could compute and compare the recall for two groups: cases corresponding to vessels flagged to Asian flag States vs. those flagged to States in other regions. 
We also compared the recall between the four gears, and, since the set of vessels of each gear type has a different composition in terms of flag States, we compared the recall of each combination of gear and flag region. 
It would have been better to also compare specificity between groups, but we only had negative cases for non-Asian trawlers. 

\begin{table}[ht]
  \caption{Number of vessel-years per gear and flag region}
  \label{table:groups}
  \centering
  \begin{tabular}{llrrr}
    \hline
    Gear & Region & Positive & Negative & Unlabeled\\
    \hline
    \multirow{2}*{Longlines} & Asia & 30 & 0 & 12850\\
    & Others & 10 & 0 & 8037 \\
    \multirow{2}*{Squid jiggers} & Asia & 14 & 0 & 5221 \\
    & Others & 0 & 0 & 801 \\
    \multirow{2}*{Purse seiners} & Asia & 6 & 0 & 1078\\
    & Others & 1 & 0 & 4232 \\
    \multirow{2}*{Trawlers} & Asia & 9 & 0 & 6150 \\
    & Others & 2 & 53 & 68887 \\
    \hline
  \end{tabular}
\end{table}

\subsection*{Transparency and reproducibility}
The codes are publicly available at \url{https://doi.org/10.5281/zenodo.7603788}. 
The training dataset is included in the repository to allow users to reproduce most of the results. 
Identity data can be sensitive information, so the vessels in the dataset have been anonymized and the flag State of each vessel is not provided.
As the negative cases cannot be publicly shared due to a confidentiality agreement, this is the only result---from the Results section---that cannot be reproduced. 
Auxiliary codes for data preprocessing, sensitivity and calibration analyses (in Supp. Mat.), and port visits, which cannot be reproduced because they would require sensitive data or negative cases, are also available in the repository.

\section*{Results}

The estimated $\alpha^*$, the upper bound of the number of positive cases within the unlabeled, was $0.28$. Using that $\alpha^*$, $28\%$ of the cases were classified as positives. This percentage varied by gear: $93\%$ for squid jiggers, $79\%$ for longliners, $63\%$ for purse seiners, and $6\%$ for trawlers  (Table \ref{table:predictionsgear}). 
A majority of the predictions had high confidence levels ($94\%$ showed more than $80\%$ confidence).
The model especially lacked confidence with squid jiggers that were classified as negative; only $54\%$ of the squid jiggers classed as negative had $>80\%$ confidence. 
There was higher confidence in the positive and negative predictions for the other gears, ranging from $69\%$ to $98\%$ of predictions in each group with more than $80\%$ confidence (Table \ref{table:confidencegear}).

\begin{table}[ht]
  \caption{Positive and negative predictions for each gear.}
  \label{table:predictionsgear}
  \centering
  \begin{tabular}{lrrrrr}
    \hline
     & Longliners & Squid jiggers & Trawlers & Purse seiners & Total \\
    \hline
    Positive & 16578 (79.2\%) & 5607 (92.9\%) & 4214 (5.6\%) & 3335 (62.7\%) & 29734 (27.7\%) \\
    Negative & 4349 (20.8\%) & 429 (7.1\%) & 70887 (94.4\%) & 1982 (37.3\%)  & 77647 (72.3\%) \\
    Total & 20927 & 6036 & 75101 & 5317 & 107381 \\
    \hline
  \end{tabular}
\end{table}

\begin{table}[ht]
  \caption{Positive and negative predictions by gear with more than $80\%$ confidence. In parentheses, the percentage of predictions with more than $80\%$ confidence from the total of predictions in that category.}
  \label{table:confidencegear}
  \centering
  \begin{tabular}{lrrrrr}
    \hline
     & Longliners & Squid jiggers & Trawlers & Purse seiners & Total \\
    \hline
    Positive & 15297 (92.3\%) & 5305 (94.6\%) & 2920 (69.3\%) & 2953 (88.5\%) & 26475 (89.0\%) \\
    Negative & 3359 (77.2\%) & 231 (53.8\%) & 69445 (98.0\%) & 1824 (92.0\%)  & 74859 (96.4\%) \\
    \hline
  \end{tabular}
\end{table}

The global recall across all vessel-years (proportion of correctly identified positive cases) was $0.89$. That means that the model correctly classified reported positives as positives 89\% of the time. For (non-Asian flagged) trawlers, the model correctly classified negative cases 98\% of the time (specificity equal to $0.98$). 

When comparing recall metrics for vessels flagged to States in Asia vs. others, the values were similar ($0.88$ and $0.92$, respectively).
The recall was high for all gears except trawlers ($0.46$) (Table \ref{table:fairness}).
Regarding combinations of gear and flag region, most combinations produced high recall ($0.88-1.00$), except for trawlers flagged to Asian and non-Asian States ($0.44$ and $0.50$, respectively), and Asian-flagged purse seiners ($0.67$). 
There were no reported cases of squid jiggers flagged to non-Asian States, so there was no data to compare.
It should be noted that the number of observed cases per combination was very small in general (Table \ref{table:fairness}).

\begin{table}[ht]
  \caption{Recall for all categories of gear and regions assessed for fairness. The number of positive cases is between parentheses.}
  \label{table:fairness}
  \centering
  \begin{tabular}{lrrr}
    \hline
     & Asia & All others & Global \\
    \hline
    Longliners & 1.00 (30) & 1.00 (10) & 1.00 (40) \\
    Squid jiggers & 1.00 (14) & - & 1.00 (14) \\
    Trawlers & 0.44 (9) & 0.50 (2) & 0.46 (11) \\
    Purse seiners & 0.67 (6) & 1.00 (1) & 0.71 (7) \\
    All gears & 0.88 (59) & 0.92 (13) & 0.89 (72) \\
    \hline
  \end{tabular}
\end{table}

\section*{Discussion}

For a model and its results to be trusted, the model should be fully detailed, the strength of its assumptions stated, and the methodological choices explained and discussed \cite{Mitchell2019}.
The algorithms should be available for inspection, ideally along with the training data, allowing for reproducibility. 
The codes should be easy to navigate with a read-me file, include comments that facilitate navigation, and should be runnable in open software \cite{Lowndes2017}.
The results should be discussed and contrasted with previous knowledge and the literature.
Confidence or uncertainty measures for the predictions should be provided \cite{Bhatt2021,Kompa2021}. 
The overall performance of the model and its fairness should be assessed, and its limitations discussed \cite{barocas-hardt-narayanan}.
In this section, we discuss each of these elements, as well as the remaining algorithmic and---more importantly---policy challenges to combat forced labor.

\subsection*{Methodological transparency}

In this work, we provide some evidence that the model can be trusted through our methodological transparency. 
The data has been fully described, including describing data limitations in terms of both size and diversity and how those limitations could lead to a biased training of the model. 
The data's characteristics conditioned the modeling approach (machine learning, positive-unlabeled, cross-validation splits accounting for sources of information to avoid data leakage, features obtained from AIS, threshold calibration through a density-based approach) which was described in detail. 
To assess the consequences of some parameterization choices, a sensitivity analysis was performed. 
The training data for the model and all model code is publicly available at \url{https://doi.org/10.5281/zenodo.7603788}.
The code was written in the R open and free software computing language \cite{R} and can be run on any computer with sufficient RAM. The code has also been optimized to take advantage of parallel processing, thus speeding up the run-time of the analysis if the computer has multiple processing cores available.
Publicly sharing the data and the code makes our methods transparent and our results reproducible. 
Additionally, the codes have been tested on different computers by different users and only slight differences were found in the predictions, due to the different ways in which Mac and Windows/Linux manage random number generators internally when calling the \texttt{ranger} package in R \cite{Rranger} (results from Linux are shown in this manuscript, while results from Mac are shown in the GitHub repository). 

\subsection*{PU learning}

The main methodological challenge was the PU learning context, which was necessary due to the near-impossibility of obtaining negative cases and lack of systematic inspections of vessels at ports---and even less at sea---concerning human rights and labor conditions.
To disentangle positives from negatives with a training sample composed of few positives and many more unlabeled cases, it is common to train classifiers that work under two assumptions. 
The first one is that the positives in the sample are representative of the positives in the population (known as the selected completely at random or SCAR assumption) and, the second one is that the unlabeled are negative \cite{Bekker2020}---in some cases, the latter assumption is relaxed by adjusting the predictions using a prior on being positive. 
To escape from the SCAR assumption, we would need to understand the sources of bias, be able to measure them, and introduce them in the algorithm for bias correction. Exhaustive quantification of bias in forced labor data collection was out of the scope of this work but could contribute to future improvements of the model.
Regarding the second assumption, \cite{Ivanov2020} developed the density-based algorithm described in the methods section and showed that it is possible to train any classifier to distinguish positives from unlabeled and then use $\alpha^*$ to reclassify the unlabeled scores into positives and negatives. 
To estimate $\alpha$ directly instead of its upper bound would have required additional assumptions such as non-overlapping distributions between positives and negatives, a subdomain of local certainty for positives, or that at least the negative distribution is not a mixture that contains the positive distribution \cite{Bekker2020}. We chose not to make any of these assumptions, as they were likely to be false in our case.
Making our predictions $\alpha^*$-dependent might result in an overestimation of the number of positives and this should be taken into consideration when interpreting the results. 

\subsection*{Positive predictions}

With the exception of trawlers (recall of $46\%$), the percentage of positives predicted by the model was high for all fishing gears ($93\%$ for squid jiggers, $79\%$ for longliners, and $63\%$ for purse seiners).
Our results are not so different from the maximum percentage predicted in \cite{McDonald2021}: $94\%$ for squid jiggers, $60\%$ for longliners, and $4\%$ for trawlers (the analysis in \cite{McDonald2021} did not include purse seiners)—the results in \cite{Karthikeyan2022} did not include predictions. 
Considering the improvements in the methodology and the increase in the number of positive cases to train the model, this consistency in a high percentage of positive cases in longliners and squid jiggers could be telling of a large forced labor problem in these fleets. 

Squid jiggers, longliners, and purse seiners are more likely to operate in the high seas than trawlers, and they all have a much higher predicted risk of forced labor than trawlers. That may not be surprising given what we know about high-seas fishing.  
In a study on the economics of high-seas fishing fleets, \cite{Sala2018} estimated that without low labor costs and government subsidies, most of the exploited high-seas fishing grounds would be unprofitable. 
Unfair labor compensation, or no compensation at all, could mean that fisheries estimated to be unprofitable are actually economically viable. 
In addition, high-seas trips are usually long, making fishing crews more vulnerable to abusive work practices and inhumane living conditions, as well as physical and psychological violence \cite{Marschke2016, Vandergeest2021a}.

Trawlers' fishing effort (even in the high seas) is generally constrained to nearshore regions within the continental shelves \cite{Taconet2019}. 
Operating closer to the coast and more frequently visiting port might reduce the chances for human rights abuses to occur by increasing the opportunities for vessel inspection (in-port or at-sea), fishing crew changes, or escaping from abusive or threatening situations before they reach the level of forced labor. 
However, although the risk may be lower, there is evidence of forced labor on trawlers: positive cases collected for this work, documentation of human rights abuses in bottom trawlers \cite{Steadman2021}, and, in a survey-based study, trawling gear was identified as an important predictor of a port being risky to receive vessels with forced labor abuse \cite{Selig2022}. 

\subsection*{Confidence levels}

An innovative component of our model is the computation of a confidence level ranging from 0 to 1 for each prediction or classification. 
We used the multiple predicted scores for each case---that resulted from using several initial random seeds and bags---to fit a Beta distribution and 
estimated confidence as the probability that the true score is above (if predicted positive, or below if predictive negative) the classification threshold.
Essentially, if all scores are far from the threshold (either high or low), the model shows consistency and high confidence in its prediction. 
But if the scores differ, the distribution will be flat and the confidence low.

A majority ($94\%$) of the predictions had high confidence ($>0.8$ confidence).
Almost all combinations of gear and class (positive/negative) showed a high percentage of predictions with high confidence.
The model was less confident in classifying a large percentage of squid jigger cases as negatives ($46\%$ had less than 0.8 confidence), so even more than the $93\%$ of squid jigger cases classified as positives may be associated with forced labor. 

\subsection*{Performance assessment}

While a model's confidence in its predictions is desirable and useful to interpret the predictions, it does not guarantee that the model is correct. 
Metrics such as recall and specificity are commonly used to measure the model's prediction performance. 
With our cross-validation procedure, we made sure that the data used to compute recall was not used to train the model. 
The overall recall was high ($0.89$), suggesting we could trust the global and average ability of the model to identify known (reported) cases of forced labor.
However, a high overall recall might hide a poor performance for a group of individuals or vessel-years. 
The recall was the only fairness metric that could be computed for all gear and flag-region groups and used to assess if (and to what extent) the model is underperforming for some of these groups. 

At the gear level, the recall was very high for longliners and squid jiggers, relatively high for purse-seiners, and low for trawlers (Table \ref{table:fairness}). 
The trawling sector is characterized by high diversity, targeting multiple species, at different depths, using different types of nets, and comprising vessels of a wide range of sizes \cite{Eigaard2011, Holland1999, Suuronen2020}. 
This heterogeneity adds a level of complexity to correctly identifying positives in this fleet. 
While difficult, subsetting the existing trawler category based on these characteristics and training the model on positive cases from each subcategory could help increase its performance in all or at least some subcategories of trawlers. 

Even though forced labor is a global problem in fisheries, there has been a significant focus on Asian fleets \cite{ILO2013}. 
The positive cases in this study reflect this disparity: $59$ ($82\%$) corresponded to vessels flagged to Asian countries. 
We assessed if the high representation of Asian-flagged cases in the positive set resulted in a high recall for cases involving vessels flagged to Asian countries and a low recall for cases involving vessels flagged in other regions. This would represent evidence of bias in the predictions against vessels flagged to Asian States. 
Overall, the model showed high recall for cases of vessels flagged to both Asian and non-Asian States and a slightly higher recall for cases involving vessels flagged non-Asian States. 
The model's poor ability to correctly identify positives in trawlers was true for Asian and non-Asian flagged cases. 
Because we did not have enough data to assess fairness for flag States directly, we did not present any statistics for flag States based on our predictions. 

Since the threshold for classifying scores into positives (or negatives) depends on $\alpha^*$, the probability of being positive is likely overestimated by the model.
While this is advantageous for the recall, it can be detrimental to the specificity. 
We only had non-Asian-flagged trawlers as negative cases to compute specificity. 
For this group, the specificity was $0.98$, meaning the model very accurately classified these as negative. 
We would need negative cases for more gear and flag-region groups to assess the model's ability to accurately identify negative cases in every group. 

\subsection*{AIS and non-broadcasting vessels}

In general, more positive and negative cases---and more diverse cases---are needed to train and evaluate the performance of the model.
An important limitation of our approach is that it relies on AIS-based features to train the model. 
Only $72$ out of $358$ collected reported cases for this work (20\%) were matched to AIS. 
While AIS is one of the largest public monitoring systems worldwide \cite{Taconet2019}, not all fishing vessels use AIS. 
Many non-broadcasting vessels have been associated with illicit activities \cite{Kroodsma2022, Oozeki2018, Park2020}, and illicit activities have been related to forced labor and human trafficking \cite{Couper2015, DeConing2011}. 
It is thus likely that the forced labor problem in fisheries is bigger than what has been revealed by our model. 
Recent studies have explored the combination of AIS with other satellite technologies like vessel monitoring systems, satellite synthetic aperture radar, visible infrared imaging radiometer suite, and high-resolution optical imagery to better assess human activity at sea \cite{Kroodsma2022, Park2020}. 
Data fusion of these different sources to produce features for this model would allow a more exhaustive assessment of forced labor at sea but would require further investigation.

\subsection*{Model interpretation}

In this work, we do not present any associations between the features used and the predicted classes for a number of reasons. 
First, while there are multiple methods available in the literature to help interpret machine learning algorithms globally (the entire model behavior, e.g. permutation feature importance, partial dependence plot) and locally (explain individual predictions, e.g. local surrogate models, Shapley values) that search for associations with the features used as inputs in the algorithm \cite{Molnar2020a}, all interpretations are based on simplifying assumptions that are rarely checked and difficult to overcome when violated \cite{Molnar2020, Rudin2019}.
Second, \cite{Krishnan2020} argued that interpretability, explainability, and transparency have not been given a definition from an epistemological point of view, and have been ill-defined using other interpretability-like words (e.g. understanding, intuitive), making it difficult or even impossible to determine whether any list of technical criteria could render an algorithm interpretable. 

Even more crucial than how to interpret a model is the question of what exactly we look for with an interpretation tool and if we need it to achieve that objective.
Researchers often use models in the hope of inferring properties or generating hypotheses about the natural world \cite{Lipton2018}, as if predictions and their relationships with features would reveal causation \cite{barocas-hardt-narayanan}.
Nonetheless, most classification models---including ours---are not designed to reflect causal relationships but are rather created to find associations between features that would optimize an accuracy rate or intra-class homogeneity. 
There is certainly a large difference between real-life forced labor and the architecture of our model.
The features used in the model themselves are not directly related to the economic, social, or political causes of forced labor \cite{Barrientos2013, Marschke2016, Stringer2016, Vandergeest2021a} but were computed because there was available AIS data to do so for the majority of fishing vessels in the world. 
The only assumption made in this case is that these features could provide enough information for an algorithm to develop multiple non-linear, non-parametric, and likely intricate classification rules to identify positive cases of forced labor. 
These relationships are numerous and complex; simplifying them for interpretation purposes would likely provide an incomplete and inaccurate representation of them, thus leading to misinterpretation \cite{Rudin2019}. 

Another purpose behind interpretability is usually trustability \cite{Lipton2018}. 
Interpretation tools might not be necessary nor sufficient for that purpose, particularly when relationships between features are as complicated as described above.
We cannot guarantee a complete understanding of the model by external researchers, but we did our best by providing a thorough description of the model and discussing its strengths and limitations regarding predictive performance and fairness. 

\subsection*{Uses of the model}

Because we did not design a causality model and did not aim to identify or assess risk factors for forced labor, the model should not be used to provide risk assessment (i.e. quantifying the increase or decrease in the chance of forced labor if vessel-level features, such as time spent at sea, change), predict future events of forced labor, or as evidence in court. 
The model was developed to provide more information where there is currently a gap about forced labor at sea by classifying yearly historical data from fishing vessels into positive or negative, giving more visibility to the problem, and becoming a source of information for port inspectors to decide which vessels to inspect. 
Boardings and inspections by competent authorities at sea or in port can be crucial for detecting forced labor. 
However, very few port authorities and patrolling agencies are able to inspect every vessel arriving in port or operating in their waters due to capacity limitations.
The historical predictions coming from our model can be useful information to help systematically prioritize limited resources for port inspections when better information is not available and alongside national monitoring, control, and surveillance efforts. 

\subsection*{Algorithm for inspections}

Considering that our model would benefit from more data to train it, a step further from this work would be to implement a sampling algorithm that would aim to maximize the probability of identifying forced labor-related issues while simultaneously improving the global performance and fairness metrics of our model. 
Both goals could seem contradictory.
On one hand, maximizing the probability of catching offenders using our model would lead to inspecting vessels with the highest predicted scores and confidence levels.
On the other hand, improving our model performance and fairness would require collecting more diverse training data and inspecting vessels with low predicted scores or low confidence levels. 
These contrasting objectives could be balanced with a contextual multi-armed bandit scheme \cite{Langford2007}, where two different sampling stages can alternate according to a reward function: an exploratory phase to gather information (e.g. when sampling randomly) and an exploitation phase to maximize reward (e.g. catching offenders or improving a model metric). 
This algorithm would need to be flexible to account for contextual elements which may vary between port or port State inspection programs (e.g. capacity for inspection, costs of inspection, other inspection goals or motivations, existing biases, or immigration laws) to ensure realistic sampling advice that would minimize profiling risks, support port authorities, and improve our model. 

\subsection*{Combating forced labor as a joint effort}

\begin{figure}[H]
    \caption{Number of port visits in 2020 visited by vessels predicted as positive. Each point represents a port, and the size of the point is scaled to the number of port visits.}
    \label{fig:port-visit-map}
    \begin{centering}
        \includegraphics[width=1.0\textwidth]{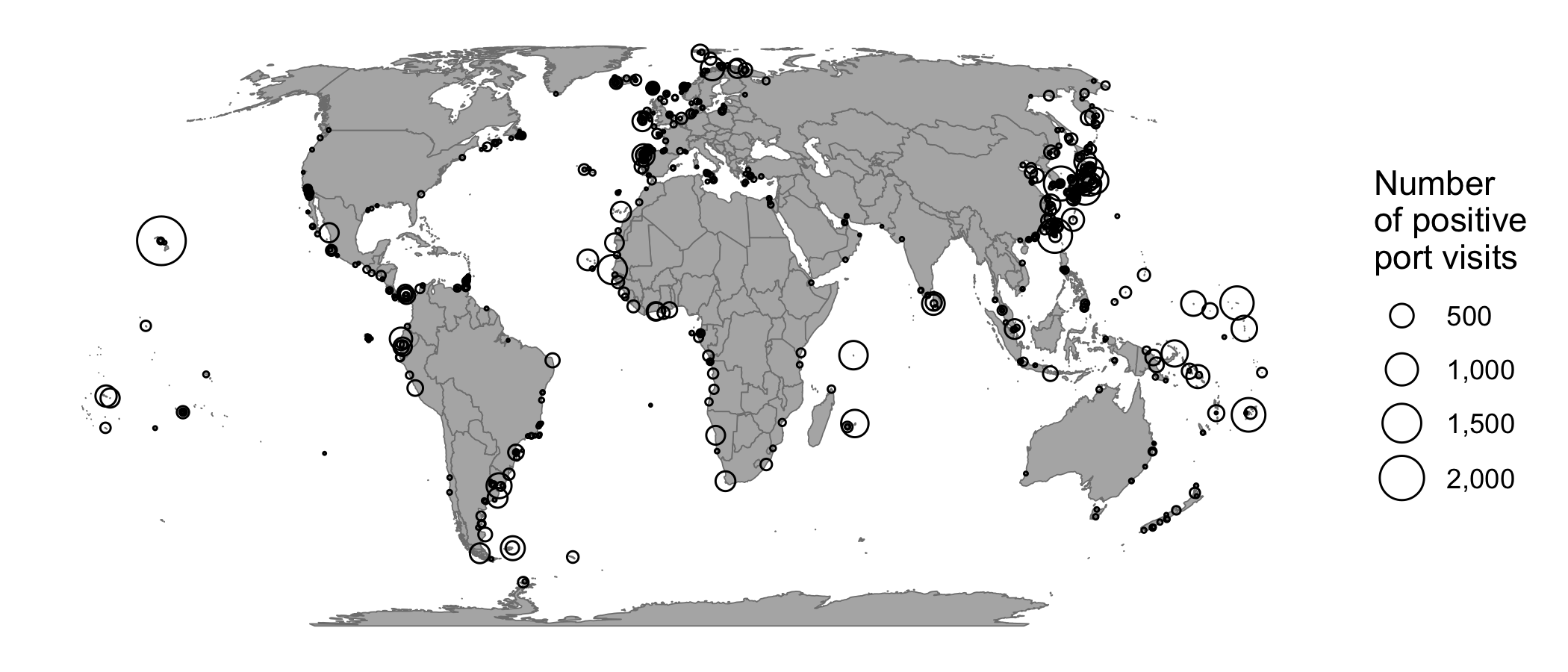}
    \par\end{centering}
\end{figure}

Estimating the prevalence of forced labor in the world and enabling more efficient port inspections are not enough to eradicate the problem. 
Vessels predicted as positive by our model visited a total of 681 ports in 2020 out of 4,248 total visited ports (16\%), across 108 countries out of 140 total countries visited (77\%) (Fig. \ref{fig:port-visit-map}).
While $76\%$ of voyages by vessels classified as positive occurred between ports within the same country, the other $24\%$ of voyages occurred between ports in different countries.
This indicates that while port- and State-level interventions will be important in addressing forced labor aboard vessels that travel within a given country, regional and cross-regional collaboration will also be necessary for vessels that travel internationally between different countries and regions.

Moreover, combating forced labor should involve all stages of labor dynamics: the point of entry for the worker, which in many cases involves recruitment brokers or agencies, the living and workplace (fishing vessels), and the point of exit for the worker (the constraints imposed on a person's ability to leave forced labor conditions, such as debt bondage or unemployment) \cite{Barrientos2013, Stringer2016}. 
It should also elevate the promotion of decent work in fisheries, not just the eradication of the most egregious types of labor abuses \cite{lozano2022decent}.

The ILO outlines that regulation of fishing can cut across many agencies. 
It is not unusual for aspects of fishing vessel registration, vessel safety, fishing licenses, catch documentation, worker safety, wages, and working conditions---to be spread across half-a-dozen or more agencies. 
It is therefore crucial that standard operating procedures exist to improve the transparency of data and ensure the timely sharing of information between these agencies, 
and face the problem through a multilateral, cooperative, and coordinated approach \cite{Ridings2021} linking flag \cite{FAO2015}, coastal, port \cite{UNCLOS}, and market States, fishers' home States, RFMOs, intergovernmental organizations, the seafood industry \cite{UNGlobalCompactSustainableSeaffod, UNGlobalCompactTenPrinciples} and civil society, and importantly include the perspectives of fishing crew members \cite{Nakamura2022}. 

The analysis of tracking data is recognized by the ILO as one of many mechanisms a competent authority may use to inform their risk assessment processes \cite{ILO2020}. 
Our research supports this recommendation. 
However, to be most helpful, tracking data-based models need to be continually updated to include a wider range of vessel gear types and inspection results.  
In this way, transparency of vessel identification, tracking, and inspection data can together contribute to a global, cost-effective tool that can support multilateral efforts to improve labor conditions at sea.

Our methods address many of the issues in \cite{McDonald2021} and \cite{Karthikeyan2022} by improving many technical aspects of the models, increasing the positive training data, and including negative cases to validate the models. 
The findings are broadly similar to \cite{McDonald2021}: the number of forced labor cases in the fishing fleet is likely very high, especially in the squid and longline fleets. 
We have also made efforts to assess the fairness or disparities in our model's performance for groups of flags and gear types. 
The advances presented here give us confidence that this model has the potential to help port inspectors and others further investigate vessels to identify this horrible practice.
Hopefully, when combined with such inspections, work like this can help improve the lives of fishers at sea. 

\section*{Authors' contributions}

To list author contributions, we used the high-level contributor roles in the CRediT framework \url{http://credit.niso.org}).

\begin{itemize}
    \item RJ:  Conceptualization, Data curation, Formal analysis, Methodology, Project administration, Software, Resources, Supervision, Validation, Visualization, Writing original draft, review, and editing, 
    \item GM: Investigation, Formal analysis, Methodology, Software, Validation, Writing (review and editing)
    \item NM: Conceptualization, Data curation, Project administration, Resources, Software, Supervision, Validation, Writing (review and editing),
    \item DK: Funding acquisition, Conceptualization, Project administration, Supervision, Writing (review and editing),
    \item CF: Funding acquisition, Resources, Investigation, Project administration, Writing (review and editing)
    \item DB: Investigation,
    \item TH: Methodology, Resources, Software, Visualization
\end{itemize}

\section*{Acknowledgments}
The authors of this piece would like to thank Kamal Azmi, Huw Thomas, and John Maefiti who were part of the forced labor risk team at Global Fishing Watch. The discussions with them about forced labor, data confidentiality, vessel inspections, and the role of the fishing industry in improving labor conditions were very valuable for this work. We would also like to thank Leah Buckley and everyone who contributed to transforming the outputs of the model into a user tool for future work with ports. 
The first author would like to give special thanks to Edgar Torres, Isis Hernández-Herrera, and Kali, for hosting her in Mérida during the preparation of the first draft of this paper. Special thanks to Krizia Matthews, Lo Ko-Jung, and Wildan Ghiffary for translating the abstract to Spanish, traditional and simplified Chinese, and Indonesian, respectively, as well as Nancy De Lemos, Enrique Tuya, Floriane Cardiec, Dame Mboup, and Jaeyoon Park for reviewing and editing the translations.

\section*{Funding}
The authors of this paper gratefully acknowledge the financial support provided by Minderoo and  Walmart Foundation that enabled this research.

\newpage

\bibliographystyle{abbrv}
 

\end{document}